\pdfoutput=1

\documentclass[11pt]{article}

\usepackage[]{EMNLP2023}

\usepackage{times}
\usepackage{latexsym}
\usepackage{graphicx}

\usepackage[T1]{fontenc}

\usepackage[utf8]{inputenc}

\usepackage{microtype}

\usepackage{inconsolata}

\title{Mavericks at ArAIEval Shared Task: Towards a Safer Digital Space - Transformer Ensemble Models Tackling Deception and Persuasion}

\author{Sudeep Mangalvedhekar\thanks{~ Equal contribution}~~, 
Kshitij Deshpande$\footnotemark[1]$~, 
Yash Patwardhan$\footnotemark[1]$~, \\
{\bf Vedant Deshpande}$\footnotemark[1]$~ \and
{\bf Ravindra Murumkar}$\footnotemark[1]$~ \\
Pune Institute of Computer Technology, Pune \\
\texttt{\{sudeepm117,kshitij.deshpande7,yash23pat,vedantd41\}@gmail.com,}\\
\texttt{rbmurumkar@pict.edu}}

\begin{document}
\maketitle
\begin{abstract}
In this paper, we highlight our approach for the "Arabic AI Tasks Evaluation (ArAiEval) Shared Task 2023". We present our approaches for task 1-A and task 2-A of the shared task which focus on persuasion technique detection and disinformation detection respectively. Detection of persuasion techniques and disinformation has become imperative to avoid distortion of authentic information. The tasks use multigenre snippets of tweets and news articles for the given binary classification problem. We experiment with several transformer-based models that were pre-trained on the Arabic language. We fine-tune these state-of-the-art models on the provided dataset. Ensembling is employed to enhance the performance of the systems. 
We achieved a micro F1-score of 0.742 on task 1-A (8th rank on the leaderboard) and 0.901 on task 2-A (7th rank on the leaderboard) respectively. 

\end{abstract}

\section{Introduction}
In today's digital age, numerous platforms aid people in reaching out to the world. However, some individuals resort to disinformation and persuasion techniques to influence people, keeping in mind a certain biased agenda, which can have negative societal effects. Disinformation \cite{wardle2017information} is an intentional effort to disseminate malicious, manipulative, and misleading information for espionage. The propagation of incorrect information can be deleterious to an individual, an organization, or a nation. Hence, disinformation detection has become imperative to catch false reports and avoid social upheaval. Persuasion is the act of changing someone's convictions, views, or conduct through interaction or exchange. Persuasion techniques can be employed to propagate propaganda \cite{alam-etal-2022-overview} and influence the behavioral patterns of the targeted audience. Persuasion can be done via textual mediums such as news articles and tweets. Social media can act as a key instrument to proliferate persuasive content as well as disinformation among the masses.

With advancements in science and technology, specifically in the domain of machine learning, machines are now capable of detecting persuasion techniques as well as disinformation from the given data. However, the detection techniques have certain limitations. The tactics used to spread disinformation constantly evolve, and the sheer volume is immense. Understanding context and intent is another challenge when it comes to detecting persuasion and disinformation. Detecting and countering these instruments of influence across multiple languages and cultural contexts can be daunting.

This paper demonstrates our work on Task 1 - Persuasion Technique Detection and Task 2 - Disinformation Detection \cite{araieval:arabicnlp2023-overview}. We intend to examine whether the given multigenre textual snippets contain persuasive content in Task 1 and classify whether the given tweet \cite{mubarak2023detecting} is disinformation or not in Task 2. Our approach highlights the use of various transformer-based models for binary classification on the given Arabic data. Ensemble-based techniques have also been employed to yield better results. 

\begin{figure*}[t]
    \centering
    \includegraphics[width=\textwidth]{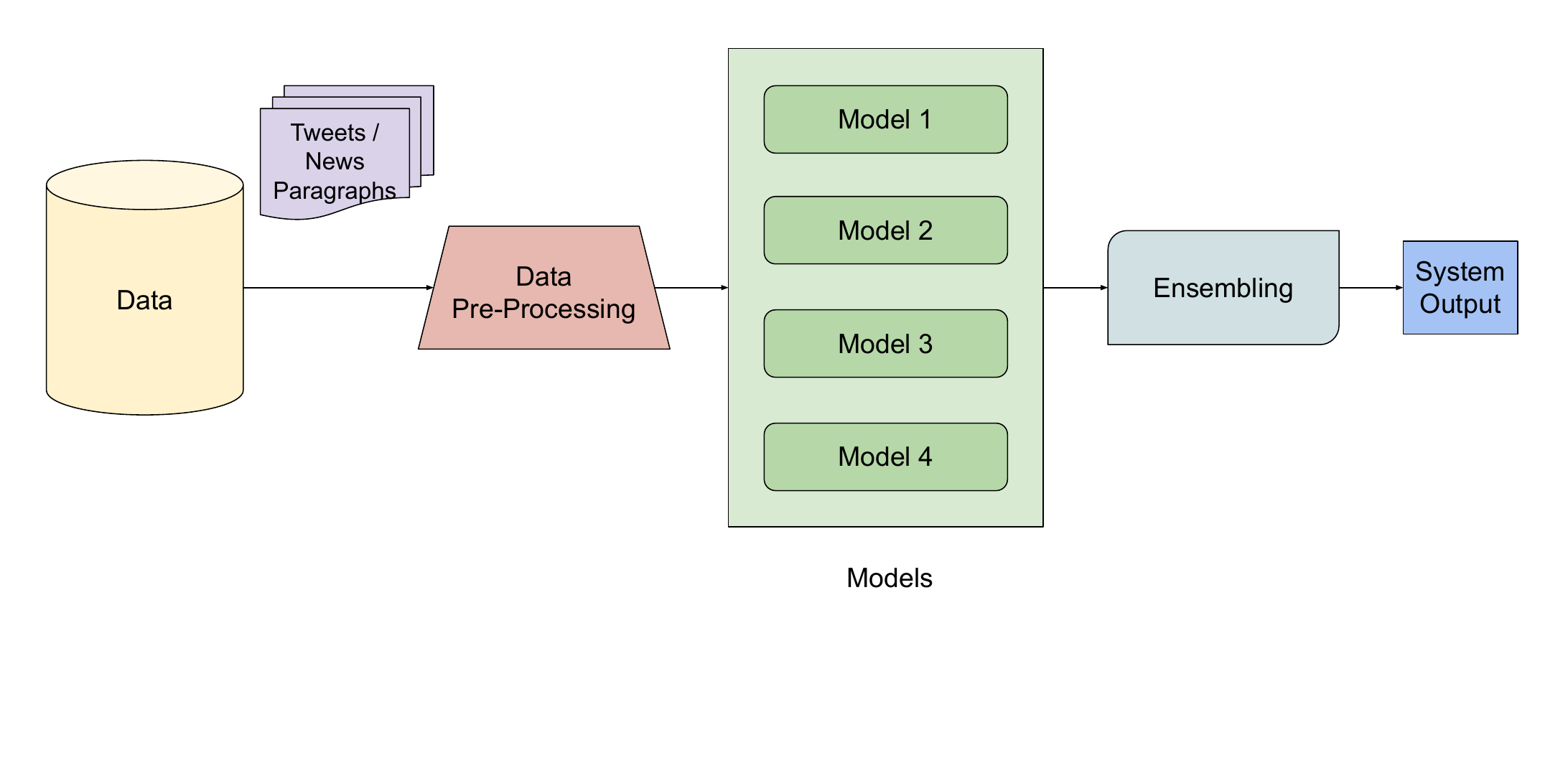}
    \caption{System architecture used for both Task 1 and Task 2}
    \label{fig:ensemble}
\end{figure*}
\section{Related Work}

In the pre-internet era, traditional media analysis, fact-checking, and investigative journalism were employed to detect disinformation and persuasion techniques. With the emergence of the internet, keyword-based approaches and sentimental analysis techniques found their groove in detecting fake news and persuading content. An analysis of linguistic features \cite{conroy2015automatic}, lexical patterns \cite{feng2012syntactic}, and rhetorical structures \cite{rubin2015towards} was used for this purpose. Further advancements in text analysis \cite{perez2017automatic} proved fruitful for this task. 

Although mitigating disinformation and persuasion techniques had been a difficult undertaking, machine learning techniques \cite{manzoor2019fake, khanam2021fake} showed promise to address this issue. Supervised machine learning methods \cite{reis2019supervised} such as Support Vector Machine(SVM), XGBoost, and Naive Bayes have been used for this purpose. \cite{iyer2019unsupervised} stated that the detection of persuasive tactics in a text can be automated using unsupervised learning. Network analysis methods \cite{shu2019studying} such as centrality measures can be used to identify coordinated behavior. This technique can help to highlight the propagation of disinformation or persuasive content over social networks.

Subsequently, deep learning techniques \cite{kumar2020fake} also contributed to fine-tuning the results. The utilization of word embeddings and convolutional neural networks (CNNs) for recognizing persuasion at an early stage also fueled the prevention of social engineering attacks \cite{tsinganos2022utilizing}. Multiple state-of-the-art systems such as LSTMs \cite{kumar2020fake} are also used to detect fake news. Further research revealed that transfer learning approaches like BERT produced more promising results than other cutting-edge NLP techniques \cite{qasim2022fine}. Ensembling techniques \cite{ahmad2020fake} were utilized to further enhance the results by integrating various approaches into a single one. Hybrid architectures like combining BERT with a recurrent neural network (RNN) \cite{kula2021application} or a combination of parallel CNNs with BERT \cite{kaliyar2021fakebert} achieved a significant score. Recent developments suggest that AI approaches such as explainable AI (XAI) \cite{chien2022xflag} are being experimented with for the task of disinformation and persuasion technique detection.

In this paper, we present our approach, which encompasses the utilization of transformer-based models for classification. Variations of BERT are used to develop an ensemble-based system for the given classification tasks.

\section{Data}

The dataset provided for Task 1 - Persuasion Technique Detection comprises multigenre text snippets, which are either tweets or news paragraphs. The training data has 2427 samples of such snippets, the development data has 259 samples, and the testing data has 503 samples. The training data contains features such as the id, text, label, and type. Each snippet in the training dataset is labeled as either 'true' or 'false' based on the presence of persuasion techniques in the given sample. This task falls under the category of binary classification.

The dataset provided for Task 2 - Disinformation Detection comprises tweets. The training data has 14147 (14126 non-null) samples of such tweets, the development data has 2111 samples, and the testing data has 3729 samples. The training data contains features such as the id, text, and label. Each tweet in the training dataset is labeled as either 'disinfo' or 'no-disinfo' based on the content in every sample. This task falls under the category of binary classification.

The provided dataset is preprocessed using regular expressions to remove irrelevant strings such as "@USER", "LINK" and "RT" to reduce the noise.
\section{System}
This shared task discusses the problems of Disinformation and Persuasion detection. These problems come under the umbrella of classification problems for which Transformer-based Models have been widely used and have achieved impressive performance. Thus, we have utilized several transformer-based models and ensembling methods in our research as shown in figure \ref{fig:ensemble}. The models are trained for 10 epochs with a learning rate of 1e-5, a batch size of 32, and the AdamW optimizer. The methodologies have been briefly discussed in the section below.
\subsection{BERT}
\citet{antoun-etal-2020-arabert} discusses how BERT models which are pre-trained on a large corpus of a specific language like Arabic, perform well on language understanding tasks. They propose several such models that help provide state-of-the-art results for the Arabic language and thus have been utilized for our research.

The pre-training dataset used for the models comprises 70 million sentences which is about 24GB in size. The data consists of news that spans multiple topics and thus represents a variety that is useful for numerous downstream tasks. The Masked Language Modeling and Next Sentence Prediction Tasks have been used as the pre-training objectives which help the models develop a good contextual understanding of the input sequence. AraBERT was evaluated on three NLP tasks namely, Question Answering, Sentiment Analysis, and Named-Entity Recognition to prove its effectiveness across various tasks and domains.

Various variants of the AraBERT model have been provided with slight tweaks in their pre-training phases and parameters used. AraBERT v1 or v0.1 are the original models, while v2 or v0.2 are the newer versions with better vocabulary and pre-processing. AraBERTv0.2-Twitter-base consists of 136M parameters, it is pre-trained with 60M multi-dialect tweets besides the dataset used for the other v0.2 models. AraBERTv2-base is pre-trained on 420M examples that have a sequence length of 128 and on 207M examples that have a sequence length of 512.

To pre-train MARBERT \cite{abdul-mageed-etal-2021-arbert}, 1B Arabic tweets were selected at random from a sizable internal dataset of roughly 6B tweets. Unlike AraBERT, the MARBERT model is trained on Twitter data, which involves both MSA and diverse dialects. It is trained using 163M parameters. This model is trained with a batch size of 256 and a maximum sequence length of 128. It is fine-tuned on several downstream tasks such as social meaning and sentiment analysis.

\subsection{ELECTRA}
Although, Masked Language Modeling pre-training for BERT-based models has given impressive results, the "Efficiently Learning an Encoder that Classifies Token Replacements Accurately" (ELECTRA) approach has yielded better results whilst being more efficient in terms of model size and compute needed for pre-training. AraELECTRA is the discriminator model (araelectra-base-discriminator) and the generator is a BERT model (araelectra-base-generator). 

The data used for pre-training consists of mostly news articles and the size of the dataset is 77GB which consists of 8.8 billion words. The model is pre-trained for 2 million steps with a batch size of 256.

AraELECTRA is a BERT-based model with 12 encoding layers consisting of 12 attention heads. Its hidden size is 768 and has a maximum input sequence length of 512. The total parameters in AraELECTRA are 136 million.
The generator Model (araelectra-base-generator) used in the ELECTRA approach for pre-training is a BERT model of a considerably smaller size with 60 million total parameters.
AraELECTRA is evaluated on three NLP tasks namely, Question Answering, Sentiment Analysis, and Named-Entity Recognition.

\section{Ensembling}
Ensembling is a technique that combines the results of various models to generate the eventual intended result of the system. Statistical as well as non-statistical methods are used for this purpose. Ensembling is useful as it helps generate results that are better than the results given by the individual models.

Amongst several methods leveraged for ensembling, we observed that the "hard voting" ensemble technique proved to be the most efficient and accurate. In hard voting, the majority vote or the "mode" of all the predictions is selected as the final prediction. It helps improve the robustness of the system and minimizes the variance in the results.


\section{Results}
We discuss the results of our experiments for tasks 1-A and 2-A in this section. Table \ref{Table:2} and Table \ref{Table:4} contain our results for the models and the ensembled score for the respective tasks. The micro F1 score serves as the official score metric for both tasks 1-A and 2-A.


\renewcommand{\arraystretch}{1.25}
\begin{table}[ht]
\centering
\begin{tabular}{|c|c|}
\hline
\textbf{Model} & \textbf{\begin{tabular}[c]{@{}c@{}}Micro F1 \\ Score\end{tabular}} \\ \hline
\textbf{Araelectra-base-discriminator} & \textbf{0.872} \\ 
AraBERTv0.2-Twitter-base & 0.842 \\ 
\textbf{MARBERTv2 (Post-evaluation)} & \textbf{0.876} \\ 
AraBERTv1-base & 0.823 \\
AraBERTv2-base & 0.849 \\ \hline
\textbf{Ensemble - Hard Voting} & \textbf{0.865} \\ \hline
\textbf{Ensemble -  Hard Voting} & \textbf{0.869} \\ 
\textbf{(Post-evaluation)} & \\ \hline
\end{tabular}
\caption{Results for Task 1-A on Development dataset}
\label{Table:1}
\end{table}

\renewcommand{\arraystretch}{1.25}
\begin{table}[ht]
\centering
\begin{tabular}{|c|c|}
\hline
\textbf{Model} & \textbf{\begin{tabular}[c]{@{}c@{}}Micro F1 \\ Score\end{tabular}} \\ \hline
\textbf{Araelectra-base-discriminator} & \textbf{0.750} \\ 
AraBERTv0.2-Twitter-base & 0.746 \\ 
MARBERTv2 (Post-evaluation) & 0.732 \\ 
AraBERTv1-base & 0.702 \\
AraBERTv2-base & 0.728 \\ \hline
\textbf{Ensemble - Hard Voting} & \textbf{0.742} \\ \hline
\textbf{Ensemble -  Hard Voting} & \textbf{0.751} \\ 
\textbf{(Post-evaluation)} & \\ \hline

\end{tabular}
\caption{Results for Task 1-A on Test dataset}
\label{Table:2}
\end{table}

\subsection{Task 1-A}
Araelectra-base-discriminator performs best with a micro F1 score of 0.872 on the development dataset and 0.750 on the test dataset as seen in Table \ref{Table:1} and Table \ref{Table:2} respectively. This performance is indicative of the advantages of utilizing the ELECTRA pre-training approach, where the Replaced Token Detection (RTD) is the objective for pre-training. It achieves a marginally better micro F1 score than the hard voting-based ensembled result of the four models. Despite this, we use the ensemble-based system as our final approach because it generates low-variance results and provides stable predictions. Our system achieved a micro F1 score of 0.742 on the test dataset.

In the post-evaluation phase (after submission of the official scores), out of the various models we experiment with for the given task, MARBERTv2 outperforms Araelectra-base-discriminator and emerges as the best model with a micro F1 score of 0.876 on development dataset. This can be attributed to the large size of the tweet-based training corpus. It boosts the ensemble scores to the 0.869 on development dataset and the 0.751 on test dataset. 
\renewcommand{\arraystretch}{1.25}
\begin{table}[ht]
\centering
\begin{tabular}{|c|c|}
\hline
\textbf{Model} & \textbf{\begin{tabular}[c]{@{}c@{}}Micro F1 \\ Score\end{tabular}} \\ \hline
Araelectra-base-generator & 0.893 \\
\textbf{AraBERTv0.2-Twitter-base} & \textbf{0.907}\\
\textbf{MARBERTv2 (Post-evaluation)} & \textbf{0.909}\\
AraBERTv1-base & 0.882 \\ 
AraBERTv2-base & 0.897 \\ 
\hline
\textbf{Ensemble - Hard Voting} & \textbf{0.909} \\ \hline
\textbf{Ensemble -  Hard Voting} & \textbf{0.914} \\ 
\textbf{(Post-evaluation)} & \\ \hline
\end{tabular}
\caption{Results for Task 2-A on Development dataset}
\label{Table:3}
\end{table}

\renewcommand{\arraystretch}{1.25}
\begin{table}[ht]
\centering
\begin{tabular}{|c|c|}
\hline
\textbf{Model} & \textbf{\begin{tabular}[c]{@{}c@{}}Micro F1 \\ Score\end{tabular}} \\ \hline
Araelectra-base-generator & 0.882 \\
\textbf{AraBERTv0.2-Twitter-base} & \textbf{0.900}\\
\textbf{MARBERTv2 (Post-evaluation)} & \textbf{0.903}\\
AraBERTv1-base & 0.882 \\ 
AraBERTv2-base & 0.894 \\ \hline
\textbf{Ensemble - Hard Voting} & \textbf{0.901} \\ \hline
\textbf{Ensemble -  Hard Voting} & \textbf{0.905} \\ 
\textbf{(Post-evaluation)} & \\ \hline
\end{tabular}
\caption{Results for Task 2-A on Test dataset}
\label{Table:4}
\end{table}

\subsection{Task 2-A}
AraBERTv0.2-Twitter-base achieves the best results with a micro F1 score of 0.907 on the development dataset and 0.900 on the test dataset as seen in Table \ref{Table:3} and Table \ref{Table:4} respectively among the four models. This is suggestive of the benefits of the model being pre-trained on a dataset consisting of tweets. The hard voting-based ensemble provides the best results as mentioned in Table \ref{Table:4}. In addition to achieving the best performance, ensembling also generates results with greater generalizability and stable predictions and is therefore chosen as the final approach for the system. Our system achieved a micro F1 score of 0.901 and a macro of F1 score 0.861 on the test dataset.

In the post-evaluation phase (after submission of the official scores), out of the various models we experiment with for the given task, MARBERTv2 outperforms AraBERTv0.2-Twitter-base and emerges as the best model with a micro F1 score of 0.909 on development dataset and 0.903 on test dataset. This can be attributed to the large size of the tweet-based training corpus. It boosts the ensemble scores to 0.914 on the development dataset and 0.905 on the test dataset. 






\section{Conclusion}
In this paper, we compared the performance of several transformer-based models on the tasks of Persuasion technique detection and Disinformation detection. For the final submission, amongst the individual models, it is observed that the Araelectra-base-discriminator achieved the best performance for Task 1-A. This model was able to achieve a micro F1 score of 0.742. Likewise, AraBERTv0.2-Twitter-base achieved the best results for Task 2-A and the final system yielded a micro F1 score of 0.901. Hard voting-based ensembling is used for our final systems to improve performance whilst also generating stable predictions. 
In the future, with the availability of better computational resources, we can enhance the system's performance by training it for longer and by using larger models. Moreover, we can experiment with other suitable ensembling techniques to gauge their effectiveness.
\label{sec:conclusion}

\section{Limitations}
Language Models used here are compute-intensive and thus may not always be suitable for application in real-world and real-time systems that have constraints on computational resources. The pre-training datasets may have certain biases in them, even though they might be rich in information. They may thus not represent the real-world picture accurately.
\label{sec:limitations}

\bibliography{anthology,custom}
\bibliographystyle{acl_natbib}

\end{document}